\documentclass[preprint,3p,twocolumn]{elsarticle}

\usepackage{amssymb}
\usepackage{amsmath}

\usepackage{hyperref}
\usepackage{multirow}
\usepackage{soul}
\usepackage{todonotes}
\usepackage{subcaption}
\usepackage{siunitx}
\usepackage{pifont}
\usepackage{numprint}
\usepackage{float}
\usepackage{tabulary}

\usepackage[cmyk]{xcolor}
\definecolor{c_unlabeled}{cmyk}{1,1,0,0}
\definecolor{c_ground}{cmyk}{1,0.44,0,0.36}
\definecolor{c_hvegetation}{cmyk}{1,0,0.64,0.27}
\definecolor{c_abutment}{cmyk}{0.89,0,1,0}
\definecolor{c_superstructure}{cmyk}{0.51,0,1,0}
\definecolor{c_deck}{cmyk}{0.14,0,1,0}
\definecolor{c_railing}{cmyk}{0,0.24,1,0}
\definecolor{c_tsign}{cmyk}{0,0.62,1,0}
\definecolor{c_pillar}{cmyk}{0,0.99,1,0}
\usepackage{tikz}

\newcommand{\cmark}{\ding{51}}%
\newcommand{\xmark}{\ding{55}}%

\newcommand\cbox[1]{%
    \tikz[baseline=-0.7ex]
        \node [minimum size=3mm, draw, fill=#1] {};}

\begin{document}

\begin{frontmatter}



\title{SemanticBridge - A Dataset for 3D Semantic Segmentation \\ of Bridges and Domain Gap Analysis}


\author[label1,label2]{Maximilian Kellner \corref{cor}}
\author[label1,label2]{Mariana Ferrandon Cervantes}
\author[label3]{Yuandong Pan \corref{cor}}
\author[label4]{Ruodan Lu}
\author[label3]{Ioannis~Brilakis}
\author[label1,label2]{Alexander Reiterer}

\affiliation[label1]{organization={Fraunhofer Institute for Physical Measurement Techniques IPM},
             city={Freiburg},
             postcode={79110},
             country={Germany}}

\affiliation[label2]{organization={University of Freiburg, Department of Sustainable Systems Engineering INATECH},
             city={Freiburg},
             postcode={79110},
             country={Germany}}

\affiliation[label3]{organization={University of Cambridge,  Department of Engineering},
             city={Cambridge},
             postcode={CB2 1PZ},
             country={United Kingdom}}
             
\affiliation[label4]{organization={Digital and Intelligent Engineering Research Institute, Sichuan Highway Planning, Survey, Design and Research Institute Ltd},
             city={Chengdu},
             postcode={610041},
             country={China}}
             
\cortext[cor]{Corresponding author.}
\begin{abstract}
We propose a novel dataset that has been specifically designed for 3D semantic segmentation of bridges and the domain gap analysis caused by varying sensors. This addresses a critical need in the field of infrastructure inspection and maintenance, which is essential for modern society. The dataset comprises high-resolution 3D scans of a diverse range of bridge structures from various countries, with detailed semantic labels provided for each. Our initial objective is to facilitate accurate and automated segmentation of bridge components, thereby advancing the structural health monitoring practice. To evaluate the effectiveness of existing 3D deep learning models on this novel dataset, we conduct a comprehensive analysis of three distinct state-of-the-art architectures. Furthermore, we present data acquired through diverse sensors to quantify the domain gap resulting from sensor variations. Our findings indicate that all architectures demonstrate robust performance on the specified task. However, the domain gap can potentially lead to a decline in the performance of up to 11.4~\% mIoU. Code and data are available at \url{https://github.com/mvg-inatech/3d_bridge_segmentation}
\end{abstract}




\end{frontmatter}

\section{Introduction}\label{ch:introduction}
Bridges are crucial for modern infrastructure and society, essential for efficient transportation and connectivity, thereby supporting economic growth by enhancing supply chain efficiency \cite{BridgesAachen0}. The state condition of this critical infrastructure is influenced by several factors. Among these are traffic volumes, structural design, aging, material type, geographic location, and environmental conditions \cite{impact_age_2001, impact_traffic_2002, impact_material_2010, impact_climate_2018, impact_traffic_2019, impact_age_2021, impact_climate_2021, impact_material_2022, impact_age_2022}.

\begin{figure}
    \centering
    \begin{subfigure}{0.45\textwidth}
        \includegraphics[width=\textwidth]{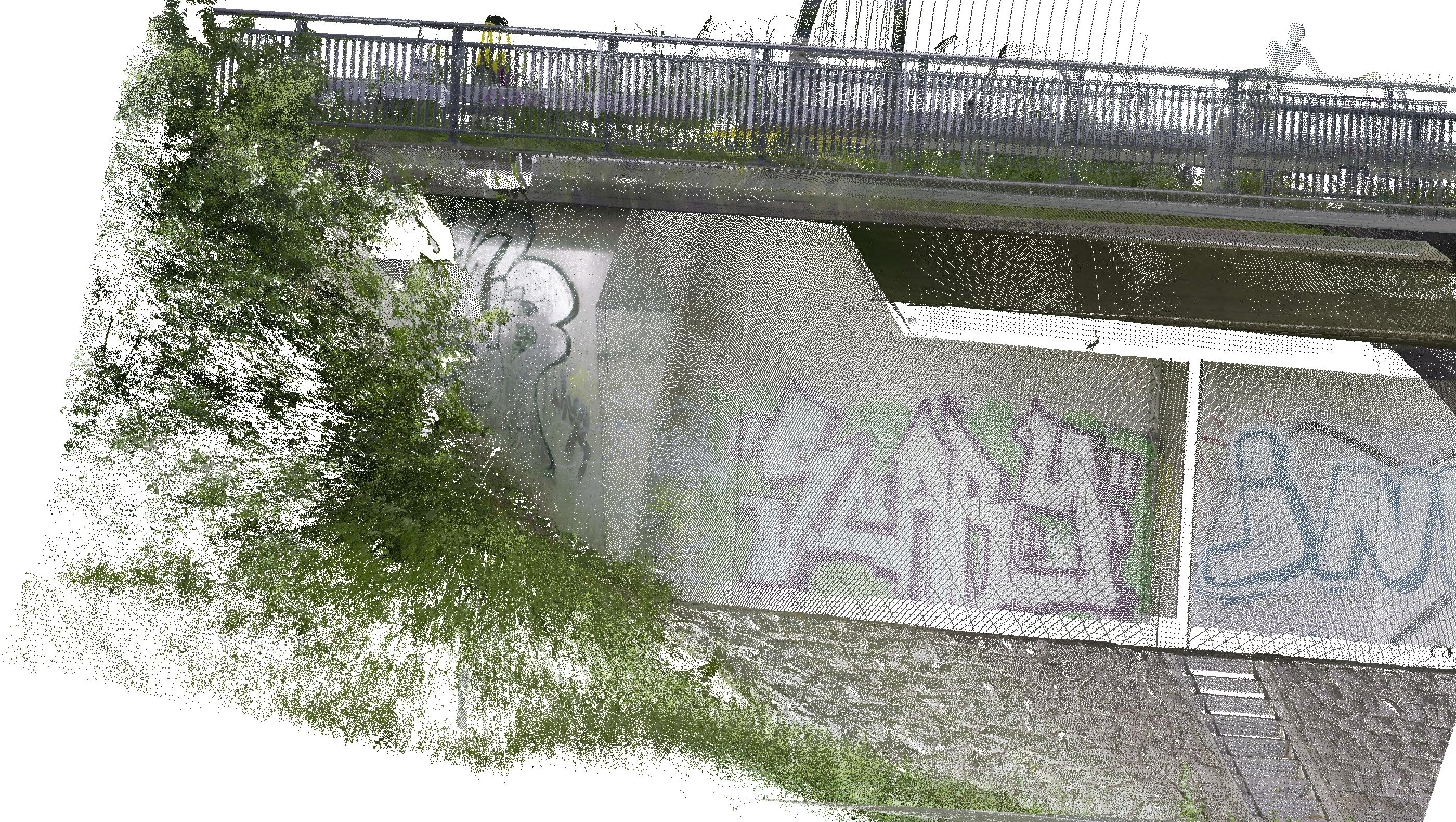}
        \caption{Example cloud from a terrestrial laser scanner.}
    \end{subfigure}
    \hfill
    \begin{subfigure}{0.45\textwidth}
        \includegraphics[width=\textwidth]{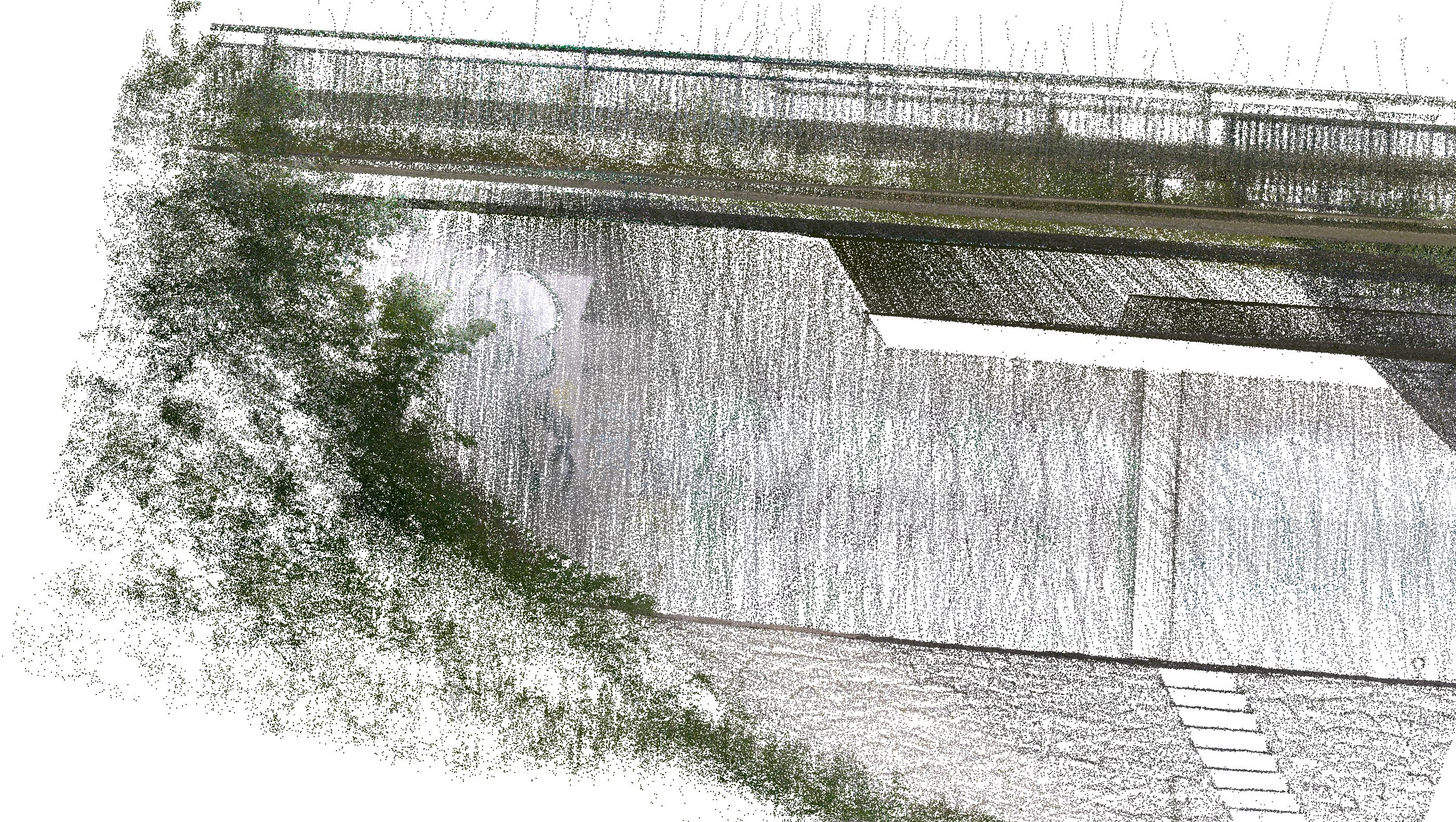}
        \caption{Example cloud from a mobile laser scanner.}
    \end{subfigure}
    \caption[]{Visualization of point clouds showing the same scene captured by two different scanners. Zoom in for a better view.}
    \label{fig:different_clouds}
\end{figure}

To ensure safe operation, these bridges are regularly inspected, with condition ratings assigned according to the observed state \cite{BMS_Asia_2017, BMS_World_2021, BMS_EUEast_2021}. The observed data are further compiled for the planning and allocation of maintenance measurements.
Responding to the need for management and maintenance of bridges, the introduction and improvement of bridge management systems has seen a steady increase in the last years \cite{BMS_2024_Fig2}. The existence of a 3D geometric model, in turn, \cite{BridgeScanningReview} could facilitate more efficient inspection procedures, simplify the process of tracking damage, and provide faster intervention options \cite{DigitalTwinHelp}. In \cite{Scan2BIM_2024} a comprehensive overview of the methodology employed in the creation of digital models of bridge assets using different data sources as input is discussed. As referenced, a common implementation of this technology is through the use of a point cloud as input and 3D semantic segmentation as the initial processing step.


3D point cloud segmentation is a computer vision technique to classify objects within a point cloud on a point-by-point basis, enriching the data by providing a detailed understanding of the content. Today, 3D segmentation is approached in a variety of ways, with a general overview provided in \cite{Survey3DLearning_2019, Survey3DLearning_2024}. Regardless of the methodology employed, all supervised methods require one common element: annotated data for training purposes. This is one significant challenge that our research aims to address. The existing 3D datasets for semantic segmentation do not address the specific challenge of segmenting scenes of bridges. The large-scale bridge components often present complex geometries whose rule-based descriptors rely on prior and usually bridge type-specific knowledge, limiting their generalization ability \cite{AutomatedBridgeComponentRecognition}. Additional challenges arise from traffic, surrounding vegetation, noise and occlusions in real-world data \cite{AutomatedSemanticSegmentationOfBridges}.

Another common problem is the so-called domain gap, which refers to the discrepancy in data distribution between the training domain and the target domain to which the model is applied. The fact that the training and test data may come from different sources, resulting in variations in data characteristics, is one of the main reasons for such a gap. For practical reasons, and especially in the case of bridge structures, this condition may be difficult to satisfy given the variety of 3D acquisition sensors on the market \cite{3DSensors}. Explicitly, we are referring to the possible variation in the recording pattern, for instance, or the sparsity and overall quality of the data. Figure \ref{fig:different_clouds} illustrates the same scene captured by two distinct sensors, which demonstrate disparate scanning patterns and resolutions. 

The objective of this paper is twofold: first, to address the previously identified challenge by introducing a new dataset for 3D semantic segmentation of bridges. This dataset comprises 20 bridges scanned with a stationary scanner in the UK and Germany, annotated across nine distinct classes. As each bridge is subjected to scanning from multiple locations, all its structural components are captured comprehensively. This facilitates research into automated surveying and reconstruction of entire bridges. To the best of our knowledge, it is the largest annotated laser-scanned point cloud dataset of bridges available. Second, this paper aims to clarify the impact of the domain gap introduced by using various complex 3D sensors. The ability to actually measure the effect that such a change has on the performance of a trained neural network is another topic that will be addressed in this paper. To this end, seven of the bridges were recorded additionally with a mobile scanner. To reduce annotation time, the annotations of the bridges scanned with the mobile device were interpolated with those of the stationary scanner. Three different 3D architectures were employed, trained, and tested on our data to identify the actual gap introduced by switching the 3D sensor.

Our experiments demonstrate that the performance of all architectures is relatively consistent when using the same sensor. However, there is a notable decline in performance when the same bridge is evaluated using a different sensor. The actual metric drop ranges from 6.9 to 11.4 \% points in terms of mean Intersection over Union (mIoU), representing a significant reduction in performance.

To summarize, our contributions with this benchmark approach are the following:
\begin{itemize}
    \item The introduction of an enriched dataset of bridges collected with Terrestrial Laser Scanners and Mobile Laser Scanners, incorporating classes such as abutments and pillars, which are not covered in previous datasets. 
    \item An investigation into the domain gap in 3D semantic segmentation, which arises from the use of disparate data capture sensors.
    \item A general comparative evaluation of the performance among three state-of-the-art architectures in the bridges semantic segmentation task.
\end{itemize}

\section{Related Work}\label{ch:related_work}
\begin{table*}[!htbp]
\footnotesize
\centering
\begin{tabular}{lcccccc}
    \hline
    Name & Sensor & Multiple devices & RGB & Scene & Points & Classes \\
    \hline
    ScanNet \cite{ScanNet} & MLS & \xmark & \cmark & Indoor & 242M & 20 \\
    S3DIS \cite{S3DIS} & TLS & \xmark & \cmark & Indoor & 215M & 13 \\
    Oakland3d \cite{Oakland3d} & MLS  & \xmark & \xmark & Road & 1.6M & 5 (44) \\
    Semantic3d \cite{Semantic3d} & TLS & \xmark & \cmark &  Road & \numprint{4000}M & 8 \\
    Paris-Lille-3D \cite{Paris-Lille-3D} & MLS  & \xmark & \xmark & Road & 143M & 9 (50) \\
    SemanticPOSS \cite{SemanticPOSS} & MLS  & \xmark & \xmark & Road & 216M & 14 \\
    SemanticKITTI \cite{SemanticKITTI} & MLS  & \xmark & \xmark & Road & \numprint{4549}M & 25 (28) \\
    nuScenes \cite{nuscenes} & MLS & \xmark & \xmark & Road & \numprint{1400}M & 32 \\
    Waymo \cite{Waymo} & MLS &  \xmark & \xmark & Road & - & 23 \\
    SynLiDAR \cite{SynLiDAR} & Synthetic & \xmark & \xmark & Road & \numprint{19482}~M & 32 \\
    DALES \cite{DALES} & MLS & \xmark & \xmark & Urban & 505M &8 (9)\\
    DublinCity \cite{DublinCity} & ALS & \xmark & \xmark & Urban & 260M & 14 \\
    ISPRS \cite{ISPRS} & MLS & \xmark & \xmark & Urban & 1.2M &9\\
    SensatUrban \cite{SensatUrban} & Photogrammetry & \xmark & \cmark & Urban & \numprint{2847}M & 13 (31)\\
    STPLS3D \cite{STPLS3D} & Synthetic & \xmark & \cmark & Urban & 150.4~M & 18 \\
    CLOI \cite{CLOI} & TLS & \xmark & \cmark & MEP & 140~M & 10 \\
    Unnamed \cite{UnnamedPipe} & TLS & \xmark & \cmark & MEP & 80~M & 5 (6) \\
    \multirow{ 2}{*}{WHU-Urban3D \cite{WHU-Urban3D}} & ALS & \xmark & \xmark & Urban & 213M & 9 \\
     & MLS & \xmark & \xmark & Road & 393M & 18 (35) \\
     MCD \cite{MCD} & MLS & \xmark & \cmark & Urban & - & 29 \\
     SynBridge \cite{3DBridgeSegmentationDA} & Synthetic & \xmark & \xmark & Bridges & 205M & 9 \\
    \hline
    SemanticBridge (ours) & TLS \& MLS & \cmark & \cmark & Bridges & 245M & 9 \\
    \hline
\end{tabular}
\caption[]{Overview of other point cloud datasets with semantic annotations. Number of classes used to evaluate and total number of classes annotated in brackets.}
\label{tab:existing_datasets}
\end{table*}

3D semantic segmentation can be divided into projection-based, discretization-based, point-based, and hybrid methods. In \cite{RangeNet}, the point cloud is projected into a spherical image and processed using 2D convolutions. Alternatively, the point cloud can be discretized into voxels and processed with 3D convolutions as demonstrated in \cite{UNet3d} and \cite{SegCloud}. The introduction of sparse 3D tensors in \cite{MinkowskiEngine} enables the efficient processing of large discretized scenes using sparse 3D convolutions. PointNet \cite{PointNet} is one of the first approaches, that directly take point clouds as input and use a special MultiLayer Perceptron (MLP) design for processing. This approach is further improved by \cite{PointNetPlusPlus} using a hierarchical structure to capture local features at multiple scales. In \cite{RandLANet} random sampling and local feature aggregation are introduced to process point clouds more efficiently. KPConv, introduced in \cite{KPConv}, defines kernel points to generalize convolutions to point clouds. PointTransformer \cite{PointTransformer} presents another method for direct point cloud processing, utilizing attention mechanisms instead of convolutions. This concept is extended in \cite{PointTransformerV2} through grouped vector attention and partition-based pooling, and further optimized in \cite{PointTransformerV3} with serialized points and an increased receptive field. Swin3D \cite{Swin3D} adapts the hierarchical vision transformers with shifted windows from \cite{SwinV2D} for 3D applications. A hybrid method combining different representations is illustrated in \cite{HybridFusion}, where spherical and bird's eye projections are fused using a KPConv layer.

Due to the large number of different 3D sensors, a considerable number of 3D datasets are currently publicly available. However, each of these datasets serves a specific research purpose and has been developed to enable the learning of defined objects in a particular environment. Since we are interested in the semantic segmentation of point clouds, we focus only on this specific objective. Furthermore, the focus of this study is not on datasets comprising solely a single object, despite the fact that these datasets serve the same objective. An overview of the available point cloud datasets, encompassing a variety of scenes and multiple objects, is provided in Table \ref{tab:existing_datasets}. The sensors employed for the purpose of data capture can be categorized as follows: mobile laser scanner (MLS), airborne laser scanner (ALS), terrestrial laser scanner (TLS), RGB-D cameras, unmanned aerial vehicle (UAV) photogrammetry or synthetic data. The following broad categories are proposed for the classification of scenes in which data were captured: roads \cite{Oakland3d, Semantic3d, Paris-Lille-3D, SemanticPOSS, SemanticKITTI, nuscenes, Waymo, SynLiDAR, WHU-Urban3D}, urban \cite{DALES, DublinCity, ISPRS, SensatUrban, STPLS3D, MCD}, indoor \cite{S3DIS, ScanNet}, and supply facilities, which are often referred to as mechanical, electrical, and plumbing (MEP) \cite{CLOI, UnnamedPipe}. Based on the scenes captured, it is clear that the majority of the data comes from autonomous driving or road detection and monitoring. The second largest category is urban scenes, followed by a small amount of indoor data and MEP data. The only publicly available 3D semantic segmentation dataset for bridges is SynBridge \cite{3DBridgeSegmentationDA}. The data was generated synthetically, and an approach was demonstrated for the utilization of domain adaptation to transfer this data to real-world scenarios. This work represents a substantial enhancement to the existing methodology, as it facilitates the integration of diverse approaches and further exploration into domain adaptation, for instance. It is possible that bridges are present in other datasets, but the relevant annotations are not provided and the necessary details are often occluded, such as the abutment due to vegetation. In order to study bridges, it is necessary to understand the specific sub-components, such as the pier or the abutment. It is clear that the existing data is insufficient to address the problem of semantically segmenting bridges.

\section{Data Capturing}\label{ch:data_capturing}
Three different scanner devices were employed in the creation of the present dataset, namely, the Focus 3D X330 scanner from FARO and the RTC360 and BLK2GO scanners from Leica Geosystems. All devices own integrated color cameras which have been used to obtain colored scan recordings. A summary of their specifications according to their corresponding technical sheets \cite{faro_datasheet, rtc360_datasheet, blk2go_datasheet} and user manuals, is presented in Table \ref{tab:sensors_specs}. As shown, the device by Faro and the RTC360 are categorized as TLS while the BLK2GO falls under the MLS category. TLS, also called stationary scanners, require a static location, usually a tripod, to acquire data in one scan session and need several scan sessions with varying locations to cover the surface to be measured. MLS devices are deemed more flexible, since they are able to capture data while moving along next to the target surface, possibly, mounted on vehicles or as in the BLK2GO case, being handheld carried by the scanner operator. As illustrated in \cite{TLSvsMLS_2022, TLSvsMLS_2024}, the MLS devices show a relevant advantage in terms of the time employed in the acquisition of data, although as observed as well in the present dataset, and as expected by the different measurement rates among the devices, the data is significantly sparser when compared to that of TLS devices. The Field-Of-View (FOV) in the Horizontal and Vertical (H/V) directions also contributes to the extent of the data recorded for each scan. The employed scanners base their operation on the Phase-Shift (PS) principle or a combination of that and the Time-Of-Flight (ToF) principle. \cite{Vosselman_2010} provides a detailed explanation of these measurement types. Briefly, under the ToF principle, a laser pulse is emitted and the distance to the surrounding surfaces is computed out of the flying duration of the pulse until it reflects back to the scanner detector. Under the PS principle, the distance is measured by the shift of the phases of an emitted continuous laser beam and its returned wave. In general, the PS principle allows for faster and more accurate measurements, although the maximum distance to the targets is smaller than that of the TOF scanners \cite{TOF_vs_PS_2020}. Apart from the measurement type, other factors play a role in the quality of the acquired data, such as the ones relative to the surface to be measured, as well as the environmental conditions in which the measurement takes place \cite{Wujanz_2018}.

\begin{table}
    \footnotesize
    \centering
    \begin{tabular}{l|r|r|r}
    \hline
         Scanner&  \begin{tabular}{@{}r@{}}Focus 3D \\ X 330\end{tabular} &  RTC360& BLK2GO\\
         Manufacturer & Faro & Leica & Leica\\
         Type & TLS & TLS & MLS \\
    \hline
        Scan range {[}m{]} & 0.6 - 330 & 0.5 - 130 & 0.5 - 25 \\
        Accuracy* {[}mm{]} & $\pm$ 2 & 1.9 & $\pm$ 3\\
        FOV(H/V) {[}°{]} & 360 / 300 & 360 / 300 & 360 / 270 \\
        Points/s & \textless 976k & \textless 2'000k & \textless 420k \\
        Principle & PS & \begin{tabular}{@{}r@{}}PS and\\ ToF\end{tabular} & \begin{tabular}{@{}r@{}}PS and\\ ToF\end{tabular}  \\
        Weight [kg] & 5.2 & 6.0 & 0.755 \\
        Release Year & 2013 & 2018 & 2022 \\
    \hline
    \end{tabular}
    \caption{Summary characteristics of the used sensors in the acquisition of data. *We refer the interested reader to the technical sheets of each sensor for the specific conditions in which the accuracy metric is determined. An in-depth discussion about the accuracy comparison among 3D scanners is found at \cite{Kersten_2022}.}
    \label{tab:sensors_specs}
\end{table}

In total, 20 different bridges were recorded. All were recorded using at least one of the stationary sensors. Seven of the bridges were recorded a second time using the mobile laser scanner. The dataset is then composed of ten bridges recorded using the TLS by Faro \cite{CambridgeData}, ten bridges recorded using the TLS by Leica and seven bridges using the MLS scanner. To make the data easier to use, we provide the following spatial information, summarized in Table \ref{tab:dataset_description}. The bridge height is measured as the height of the bounding box containing the ground and deck points. The length describes the distance from the beginning of the deck in one of the bridge sides until the end of the deck and the beginning of the ground at the opposite bridge side. The width is the average measured width of the deck at the two prior referred locations. The span refers to the distances within the deck surface among the centers of pillars. The number of pillars is considered as the disjoint instances of point clouds of the class "Pillar". Measurements are for reference only and were performed over the point clouds. Finally, the crossing describes the obstacle that is crossed by the bridge.

\begin{table*}
\footnotesize
\centering
\begin{tabular}{lrrrrrrr}
    \hline
    Bridge & Height (m) & Length (m) & Width (m) & \# spans & Spans lengths (m) & \# pillars & Crossing \\
    \hline
    bridge\_1\_cb & \multicolumn{1}{r}{7.67} & \multicolumn{1}{r}{62.64} & \multicolumn{1}{r}{11.14} & \multicolumn{1}{r}{4} & \multicolumn{1}{r}{12.58, 18.86, 19.04, 12.17} & 3 & road\\
    bridge\_2\_cb & \multicolumn{1}{r}{8.28} & \multicolumn{1}{r}{53.82} & \multicolumn{1}{r}{11.30} & \multicolumn{1}{r}{4} & \multicolumn{1}{r}{12.28, 14.95, 14.79, 11.80} & 3 & road\\
    bridge\_3\_cb & \multicolumn{1}{r}{8.64} & \multicolumn{1}{r}{52.37} & \multicolumn{1}{r}{11.68} & \multicolumn{1}{r}{4} & \multicolumn{1}{r}{11.31, 14.91, 14.80, 11.35} & 3 & road\\
    bridge\_4\_cb & \multicolumn{1}{r}{9.16} & \multicolumn{1}{r}{32.46} & \multicolumn{1}{r}{26.81} & \multicolumn{1}{r}{3} & \multicolumn{1}{r}{8.27, 14.42, 9.78} & 12 & road\\
    bridge\_5\_cb & \multicolumn{1}{r}{8.03} & \multicolumn{1}{r}{61.43} & \multicolumn{1}{r}{13.07} & \multicolumn{1}{r}{3} & \multicolumn{1}{r}{15.04, 30.77, 15.61} & 2 & road\\
    bridge\_6\_cb & \multicolumn{1}{r}{7.84} & \multicolumn{1}{r}{54.39} & \multicolumn{1}{r}{14.18} & \multicolumn{1}{r}{4} & \multicolumn{1}{r}{11.54, 15.54, 15.38, 11.93} & 6 & road\\
    bridge\_7\_cb & \multicolumn{1}{r}{8.61} & \multicolumn{1}{r}{39.91} & \multicolumn{1}{r}{11.76} & \multicolumn{1}{r}{2} & \multicolumn{1}{r}{21.76, 18.15} & 1 & road\\
    bridge\_8\_cb & \multicolumn{1}{r}{9.64} & \multicolumn{1}{r}{52.73} & \multicolumn{1}{r}{9.69}  & \multicolumn{1}{r}{4} & \multicolumn{1}{r}{12.77, 14.80, 14.71, 10.46} & 6 & road\\
    bridge\_9\_cb & \multicolumn{1}{r}{8.34} & \multicolumn{1}{r}{52.18} & \multicolumn{1}{r}{12.95} & \multicolumn{1}{r}{4} & \multicolumn{1}{r}{12.67, 14.54, 14.80, 10.17} & 6 & road\\
    bridge\_10\_cb & \multicolumn{1}{r}{11.75} & \multicolumn{1}{r}{69.59} & \multicolumn{1}{r}{13.50} & \multicolumn{1}{r}{4} & \multicolumn{1}{r}{16.07, 18.58, 18.74, 16.20} & 6 & road\\
    bridge\_11\_fr & \multicolumn{1}{r}{6.73} & \multicolumn{1}{r}{16.42} & \multicolumn{1}{r}{5.27} & \multicolumn{1}{r}{-} & \multicolumn{1}{r}{-} & 0 & road \\
    bridge\_12\_fr & \multicolumn{1}{r}{6.39} & \multicolumn{1}{r}{40.50} & \multicolumn{1}{r}{7.54} & \multicolumn{1}{r}{-} & \multicolumn{1}{r}{-} & 0 & river\\
    bridge\_13\_fr & \multicolumn{1}{r}{4.68} & \multicolumn{1}{r}{27.45} & \multicolumn{1}{r}{2.10} & \multicolumn{1}{r}{-} & \multicolumn{1}{r}{-} & 0 & river\\
    bridge\_14\_fr & \multicolumn{1}{r}{7.30} & \multicolumn{1}{r}{14.64} & \multicolumn{1}{r}{12.61} & \multicolumn{1}{r}{-} & \multicolumn{1}{r}{-} & 0 & road\\
    bridge\_15\_fr & \multicolumn{1}{r}{6.95} & \multicolumn{1}{r}{17.16} & \multicolumn{1}{r}{4.27} & \multicolumn{1}{r}{-} & \multicolumn{1}{r}{-} & 0 & road\\
    bridge\_16\_fr & \multicolumn{1}{r}{4.94} & \multicolumn{1}{r}{44.72} & \multicolumn{1}{r}{9.88} & \multicolumn{1}{r}{-} & \multicolumn{1}{r}{-} & 0 & river\\
    bridge\_17\_fr & \multicolumn{1}{r}{8.62} & \multicolumn{1}{r}{41.71} & \multicolumn{1}{r}{12.43} & \multicolumn{1}{r}{2} & \multicolumn{1}{r}{21.31, 20.41} & 1 & road \\
    bridge\_18\_fr & \multicolumn{1}{r}{8.42} & \multicolumn{1}{r}{37.95} & \multicolumn{1}{r}{13.51} & \multicolumn{1}{r}{3} & \multicolumn{1}{r}{12.61, 13.60. 11.73} & 4 & railing\\
    bridge\_19\_fr & \multicolumn{1}{r}{6.57} & \multicolumn{1}{r}{35.23} & \multicolumn{1}{r}{13.20} & \multicolumn{1}{r}{3} & \multicolumn{1}{r}{11.23, 13.11, 10.89} & 2 & railing\\
    bridge\_20\_fr & \multicolumn{1}{r}{7.65} & \multicolumn{1}{r}{40.16} & \multicolumn{1}{r}{22.89} & \multicolumn{1}{r}{3} & \multicolumn{1}{r}{12.71, 12.98, 14.46} & 2 & railing\\
    \hline
\end{tabular}
\caption[]{Spatial dataset overview.}
\label{tab:dataset_description}
\end{table*}

\section{Data Processing}\label{ch:data}
\begin{table}
\footnotesize
\centering
\begin{tabular}{lp{5cm}}
    \hline
    Class & Definition \\
    \hline
    \vtop{\hbox{\strut Abutment }\hbox{\strut \includegraphics[width=0.05\textwidth, height=4mm]{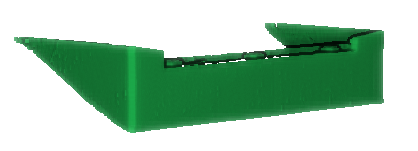}}} & The abutment is the structure that connects the deck of the bridge to the ground at the ends of a span. \\
    \vtop{\hbox{\strut Superstructure }\hbox{\strut \includegraphics[width=0.05\textwidth, height=4mm]{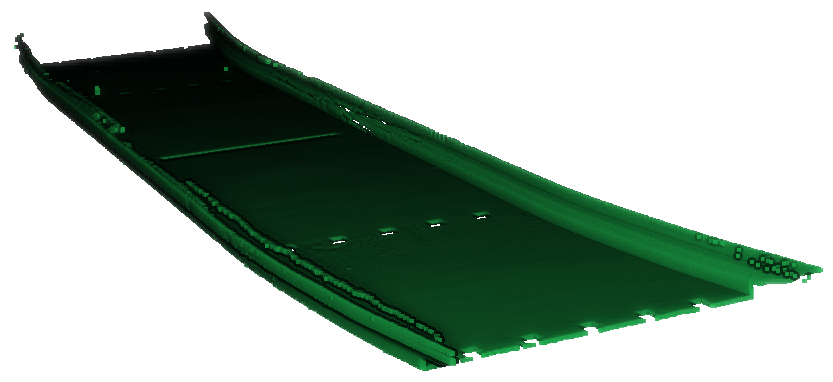}}} & The superstructure is comprised solely of the spanning elements, excluding the top surface. \\
    \vtop{\hbox{\strut Deck }\hbox{\strut \includegraphics[width=0.05\textwidth, height=4mm]{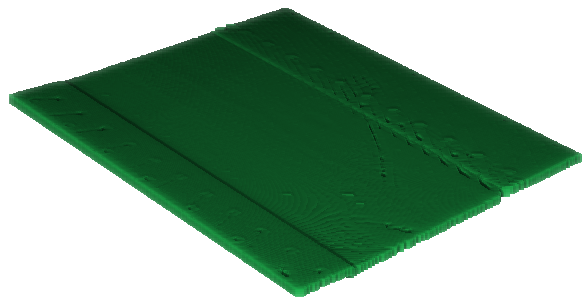}}} & The whole upper part of the bridge. This could be a road or a railway. It ends on top of the abutment. \\
    \vtop{\hbox{\strut Pillar }\hbox{\strut \includegraphics[width=0.05\textwidth, height=4mm]{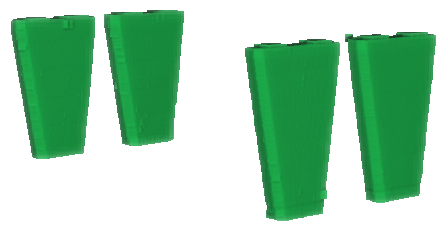}}} & The support of the superstructure without the abutment. Can have different shapes. \\
    \vtop{\hbox{\strut Railing }\hbox{\strut \includegraphics[width=0.05\textwidth, height=4mm]{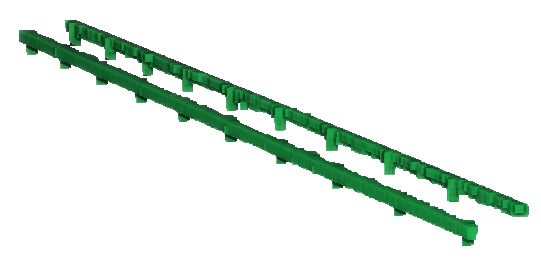}}} & All structures designed to provide support, safety and guidance for people and vehicles. This includes traffic barriers. \\
    \hline
    \vtop{\hbox{\strut High Vegetation }\hbox{\strut \includegraphics[width=0.05\textwidth, height=4mm]{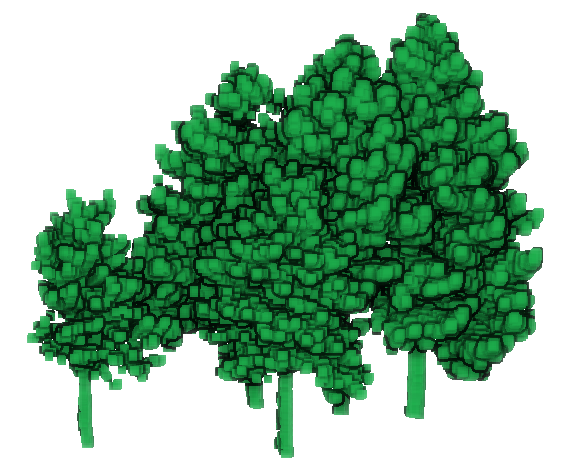}}} & Refers to bushes, shrubs, foliage and other vegetation that can be clearly identified. \\
    \vtop{\hbox{\strut Ground }\hbox{\strut \includegraphics[width=0.05\textwidth, height=4mm]{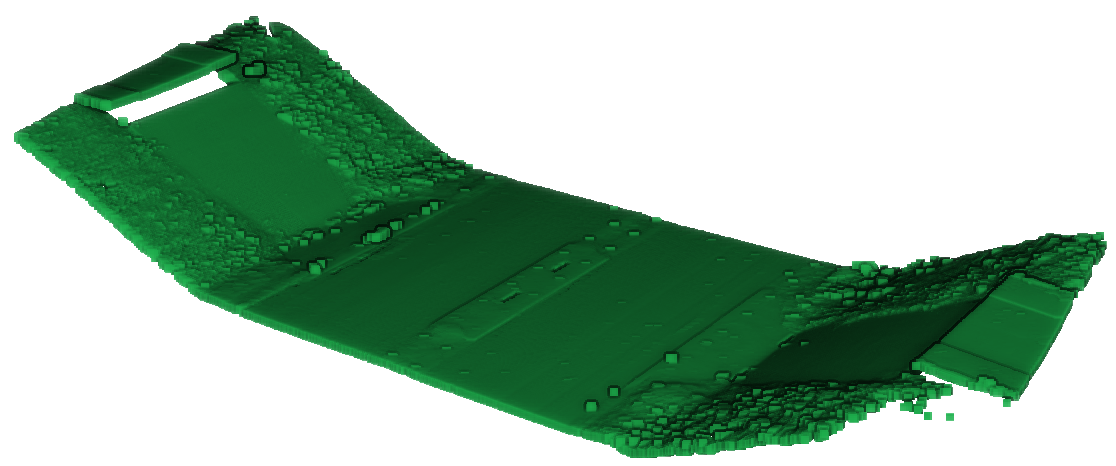}}} & The low-lying surroundings of the bridge, including roads, grass and other types of horizontally spreading vegetation, including soil. \\
    \vtop{\hbox{\strut Traffic Sign }\hbox{\strut \includegraphics[width=0.05\textwidth, height=4mm]{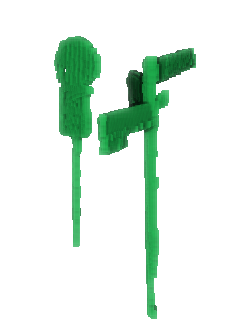}}} & Traffic signs and other objects relevant to the road system, including their mounting. This includes street lighting. \\
    \vtop{\hbox{\strut Unlabeled }\hbox{\strut \includegraphics[width=0.05\textwidth, height=4mm]{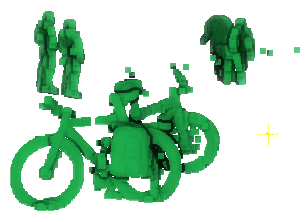}}} & All other objects that are not defined, including noise. \\
    \hline
\end{tabular}
\caption[]{Class definitions.}
\label{tab:class_definitions}
\end{table}

All data captured with stationary scanners have been manually annotated, using CloudCompare \cite{CloudCompare}. Those captured with the mobile scanner were interpolated out of the respective manual annotations instead and were only verified manually. The categories used for the annotation of this dataset were selected according to the most common bridge components, as well as the most frequently occurring objects within the recorded scenes. Accordingly, the components of the actual bridge structure are classified into the following categories: "Abutment", "Superstructure", "Deck", "Railing", and "Pillar". The surrounding environment is further divided into the following categories: "Ground", which refers to the low-level surroundings of the bridge; "High Vegetation", which refers to trees and bushes; and "Traffic Signs", which refers to signs in the environment. The class definition is summarized in Table \ref{tab:class_definitions}. This choice is also in line with the commonly found classes proposed in \cite{BridgeAnnotationRules}, where, additionally, online datasets are discussed and the absence of labeled datasets of bridges is reported. Further, the data is aligned with \cite{3DBridgeSegmentationDA} to allow incorporation of the proposed technique and combining the data together.

In order to annotate the mobile scanner points in addition to the manually classified points from the stationary scanner, we have devised an interpolation scheme. This consists of the assignment of the most frequent label within the surrounding area of each point among the aligned clouds; thereby interpolating the annotation from the manually labeled $\mathcal{P}^1$ to the unlabeled cloud $\mathcal{P}^2$. To this end, we initially conducted an alignment of the involved clouds. Afterward, the Euclidean distance between each point $\mathbf{p}_i^2$ of the point cloud $\mathcal{P}^2 \in \mathbb{R}^{N \times 3}$ for $i \in [0, N-1]\ $ and all points from point cloud $\mathcal{P}^1 \in \mathbb{R}^{M \times 3}$ with $j \in [0, M-1]\ $ is calculated:

\begin{equation}
    \mathbf{d}_i = \left\lVert  (\mathcal{P}^1 - 1_M \mathbf{p}_i^T)_j \right\rVert_2
\end{equation}

\begin{figure}
    \centering
    \begin{subfigure}{0.45\textwidth}
        \includegraphics[width=\textwidth]{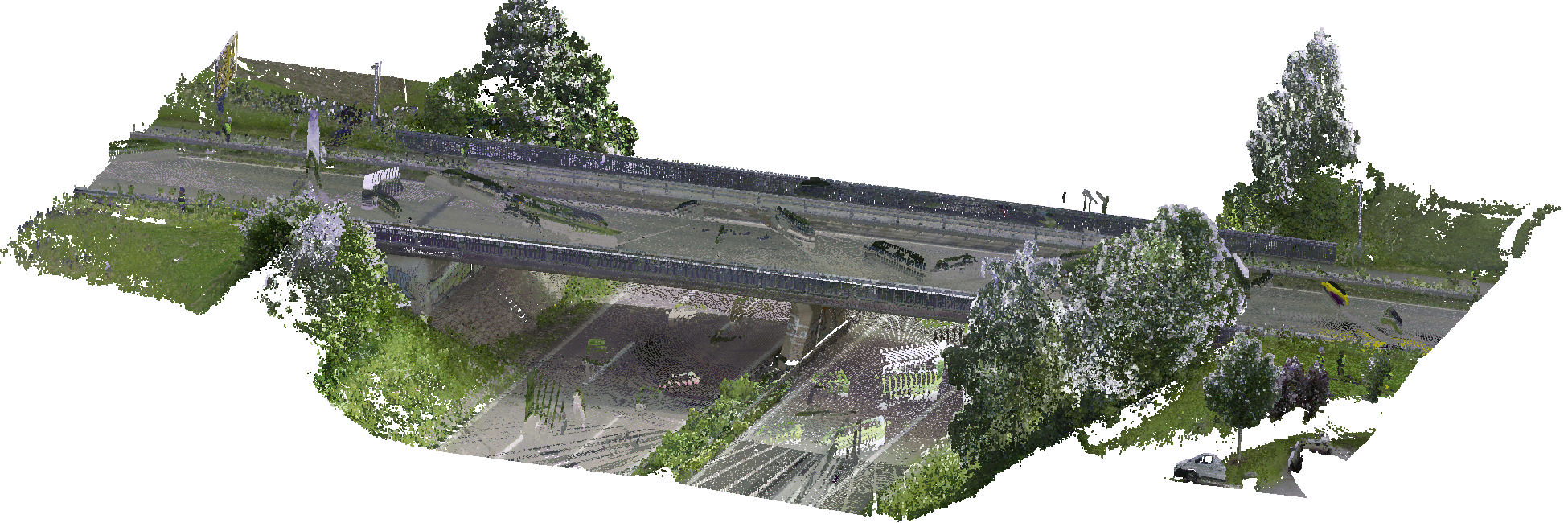}
        \caption{Colored point cloud}
    \end{subfigure}
    \hfill
    \begin{subfigure}{0.45\textwidth}
        \includegraphics[width=\textwidth]{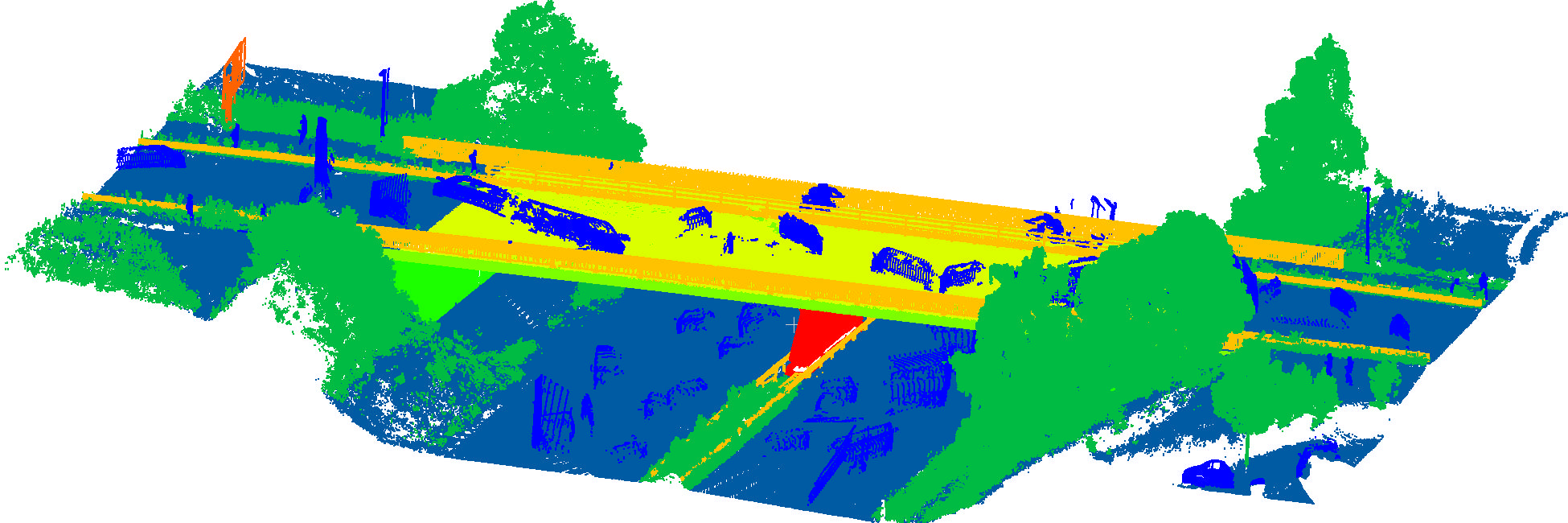}
        \caption{Manually annotated point cloud}
    \end{subfigure}
    \hfill
    \begin{subfigure}{0.45\textwidth}
        \includegraphics[width=\textwidth]{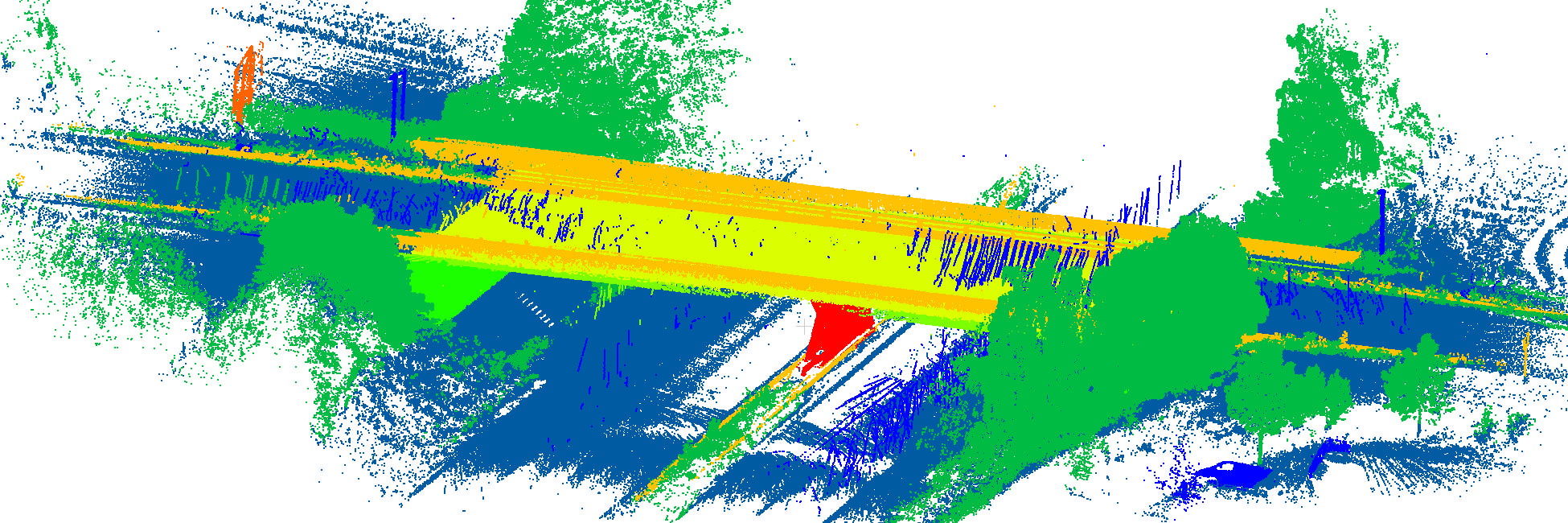}
        \caption{Automatically annotated point cloud}
    \end{subfigure}
    \hfill
    \begin{subfigure}{0.48\textwidth}
    {
        \begin{tikzpicture}
        \draw [fill=white,draw=none] (0,.5) rectangle +(8,2);
        
        \foreach \x/\c/\t/\y in {
                                 0/c_abutment/Abutment/1.8,
                                 2.2/c_deck/Deck/1.8,
                                 4.4/c_railing/Railing/1.8,
                                 0/c_pillar/Pillar/1.2,
                                 4.4/c_superstructure/Superstructure/1.2,
                                 2.2/c_unlabeled/Unlabeled/1.2,
                                 0/c_tsign/Traffic Sign/.6,
                                 2.2/c_ground/Ground/.6,
                                 4.4/c_hvegetation/High Vegetation/.6}
            {
            \fill[color=\c] (\x+0.3, \y) rectangle +(0.33, 0.33);
            \node [font=\small, anchor=west] at (\x+0.6, \y+.1) {\t}; 
            }
                
        \end{tikzpicture}
    }
    \end{subfigure}
    \caption[]{Visualization of point clouds showing a colored bridge and a bridge which was annotated manually. The last shows an example of automatic label interpolation.}
    \label{fig:example_point_clouds}
\end{figure}
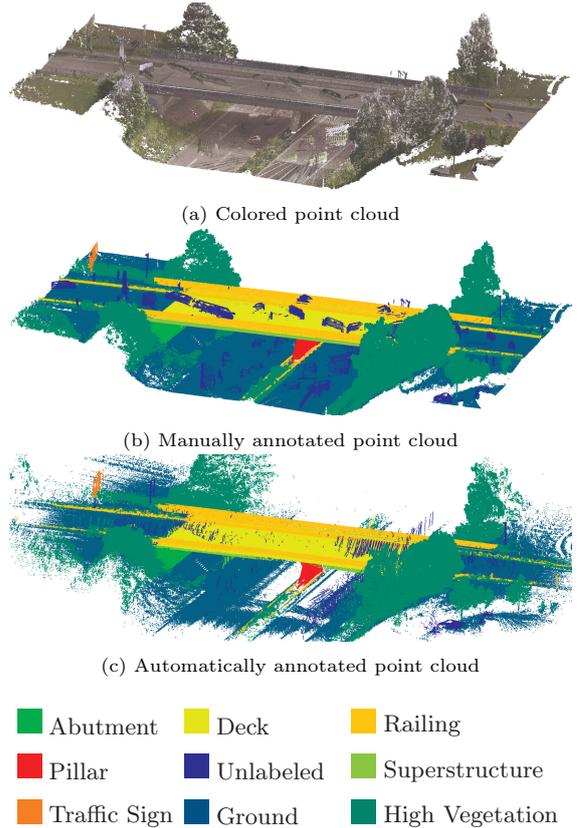

In order to determine the most probable class, the $k=8$ closest proximity to the given point is considered. The class that receives the majority of votes is then assigned to the point $\mathbf{p}_i^2$. The points with $\min(\mathbf{d}_i) > \theta$ with $\theta=0.5$ have been dropped from the point cloud in order to avoid mislabeling. Figure \ref{fig:example_point_clouds} illustrates an example of a colored point cloud, together with the manual annotations and the resulting annotations from the above-described interpolation. More examples of manually annotated bridges can be found in the Supplementary Material.

The data were then split in order to facilitate further analysis. To this end, a training split was created which exclusively uses data captured by the stationary sensors, which has a total of 15 bridges (eight clouds from the TLS by Faro and seven clouds from the TLS by Leica). This leaves five different bridges for testing (two clouds from the TLS by Faro and three clouds from the TLS by Leica). To investigate the domain gap, three of the seven bridges captured by the MLS were selected for evaluation. This selection was made to align the experiment with the dataset split setup, as the point clouds of the remaining four bridges, captured by TLS, had already been used in the training process. This will allow for further research to be conducted into whether and how these bridges could be used to reduce the domain gap. Summarized, our training and testing splits are as described in Table \ref{tab:data_splits}. The final amount of points and the classification of these points are presented in Table \ref{tab:class_list}. 

Our first priority is to provide high quality data that is easy to use, rather than the quantity of available points. For this reason, we process the data first. Given the large number of data points in the raw point clouds, and to avoid redundancy in further computations, we decided to subsample the point clouds. This was achieved by applying a voxel grid downsampling technique using a voxel size of \SI{1}{cm^3} applied to all point clouds.

\begin{table}
\footnotesize
\begin{tabular}{lll}\hline
Sensor, Ref. & Indices Train & Indices Test\\ \hline
TLS, cb\_faro  & 2-6, 8-10 & 1, 7 \\
TLS, fr\_rtc & 11, 12, 14-16, 18, 20 & 13, 17, 19 \\
MLS, fr\_blk & 12*, 15*, 16*, 18* & 13, 17, 19 \\\hline
\end{tabular}
\caption{Proposed split indices. Indices marked with * were not used within our evaluation benchmark but are made available in the dataset for further works.}
\label{tab:data_splits}
\end{table}

\begin{table*}
\centering
\begin{tabular}{lccrrrr}\hline
    Class & ID & Color & TLS train & TLS test & MLS train & MLS test \\
    \hline
    Unlabeled & 0 & \cbox{c_unlabeled} & \numprint{2935701} & \numprint{1912782} & \numprint{349887} & \numprint{366172} \\ 
    Ground & 1 & \cbox{c_ground} & \numprint{73570291} & \numprint{33314050} & \numprint{17583577} & \numprint{11634200} \\ 
    High Vegetation & 2 & \cbox{c_hvegetation} & \numprint{24137614} & \numprint{21055979} & \numprint{9446453} & \numprint{9651229} \\ 
    Abutment & 3 & \cbox{c_abutment} & \numprint{4785559} & \numprint{4208608} & \numprint{1254427} & \numprint{1321323} \\ 
    Superstructure & 4 & \cbox{c_superstructure} & \numprint{26283631} & \numprint{11563451} & \numprint{4685818} & \numprint{2972039} \\ 
    Deck & 5 & \cbox{c_deck} & \numprint{13494247} & \numprint{6352306} & \numprint{6901838} & \numprint{1926408} \\  
    Railing & 6 & \cbox{c_railing} & \numprint{7838671} & \numprint{3876508} & \numprint{2605598} & \numprint{1299639} \\ 
    Traffic Sign & 7 & \cbox{c_tsign} & \numprint{162925} & \numprint{60309} & \numprint{101674} & \numprint{53659} \\ 
    Pillar & 8 & \cbox{c_pillar} & \numprint{7735165} & \numprint{1809829} & \numprint{136856} & \numprint{260509} \\  
    \hline
    $\sum$ &  &  & \numprint{160943804} & \numprint{84153822} & \numprint{43066128} & \numprint{29485178} \\ 
    \hline
\end{tabular}
\caption{Semantic segmentation class list with respective amount of points. Splitted by proposed train and test sets.}
\label{tab:class_list}
\end{table*}

\section{Experiments}\label{ch:experiments}
In the initial experiments utilizing the proposed data, a diverse range of architectures was employed. Given our objective of remaining within the 3D domain, we have not employed projection approaches. Instead, we have selected a discretization approach utilizing a U-Net \cite{UNet} with sparse convolutions \cite{SPConv} and residual blocks \cite{ResidualNetwork} named UNet3D in the following. We have also chosen KPConv \cite{KPConv}, as it offers a generalized form of convolution for point clouds, and PointTransformerv2 \cite{PointTransformerV2} (named PTv2, in the following), as attention methods are particularly well-suited to point clouds, given their status as a set operator and the fundamental nature of point clouds as a set of points.

\subsection{Architecture Details}

We will use the point coordinates $x,y,z$ in $\mathcal{P} \in \mathbb{R}^{N \times 3}$ with $N$ points, as well as the colour information $r,g,b$ as features $\mathcal{F} \in \mathbb{R}^{N \times 3}$ as input. We have closely followed the specifications in the respective publications for selecting the parameters of the examined architectures. However, it is important to note that our objective is not to achieve the highest level of optimization of the given architectures for the purpose of segmentation. Rather, our aim is to measure and compare the discrepancy between the different 3D models in the case of a change in the sensor providing the input data. For the UNet3D model, the voxel size is set to increase by a factor of two with each layer during the encoding process. The same factor is used for the feature dimension in each layer: $32 \rightarrow 64 \rightarrow 128 \rightarrow 256$. In order to ensure that black points are also considered, the input features $\mathcal{F}$ are supplemented by a vector with a constant value of one $1_n$. The described input features, along with the corresponding increment values of the dimensions of the features, are also employed in the context of KPConv, but with a starting value of 64. With regard to pooling, the same downsampling procedure is employed as for UNet3D, whereas nearest neighbor interpolation is used for upsampling. The kernel points are set to 15, and the convolution radius grows proportionally to the subsampling grid size, beginning with $0.2$. For PTv2 the voxel size gets increased with the following multiplier  $[\times 3.0, \times 2.5, \times 2.5, \times 2.5]$. The feature dimensions $\mathcal{F}$ are combined with the point coordinates $\mathcal{P}$ and then embedded into 48 channels. These channels are further encoded into $48 \rightarrow 96 \rightarrow 192 \rightarrow 384 \rightarrow 512$ channels. The number of neighboring points used was limited to $k=16$. The four stages of encoders and decoders are employed with block depths of $[2, 2, 6, 2]$ and $[1, 1, 1, 1]$, respectively. In all cases, the initial voxel size of the input point cloud is \SI{8}{cm}.

\subsection{Implementation Details}
\label{ch:experiments_implementation}

All implementations are done using the PyTorch library \cite{PyTorch} and NumPy \cite{numpy}. We use the SGD optimizer with momentum and weight decay set to 0.9 and 0.0001, respectively. The initial learning rate is set to $0.001$, with exponential decay using the decay constant of $\gamma=0.98$. All networks are optimized by minimizing the standard cross-entropy loss using label smoothing with $\epsilon=0.1$ \cite{LabelSmoothing}.  A single NVIDIA A100 GPU is used for training, and the batch size is set to six. To further increase the data variance the following augmentations are applied to the point coordinates. First, we drop a random amount of $\mathcal{U}(0, 0.2)$ points. Afterward, the $x$, $y$, and $z$ position of each point gets shifted by the value of $\mathcal{U}(-1, 1)$, the point cloud gets rotated around the z-axis by an angle between \ang{-180} to \ang{180}, each point cloud gets scaled by a value of $\mathcal{U}(0.9, 1.1)$, we randomly flip the $x$ and $y$ axis, and we add Gaussian noise with $\mathcal{N}(\mu=0, \sigma^2=0.01)$. The color features are augmented using random drop of $\mathcal{U}(0, 0.3)$ and Gaussian noise $\mathcal{N}(\mu=0, \sigma^2=0.01)$.

\begin{figure}
    \centering
    \begin{subfigure}{0.45\textwidth}
        \includegraphics[width=\textwidth]{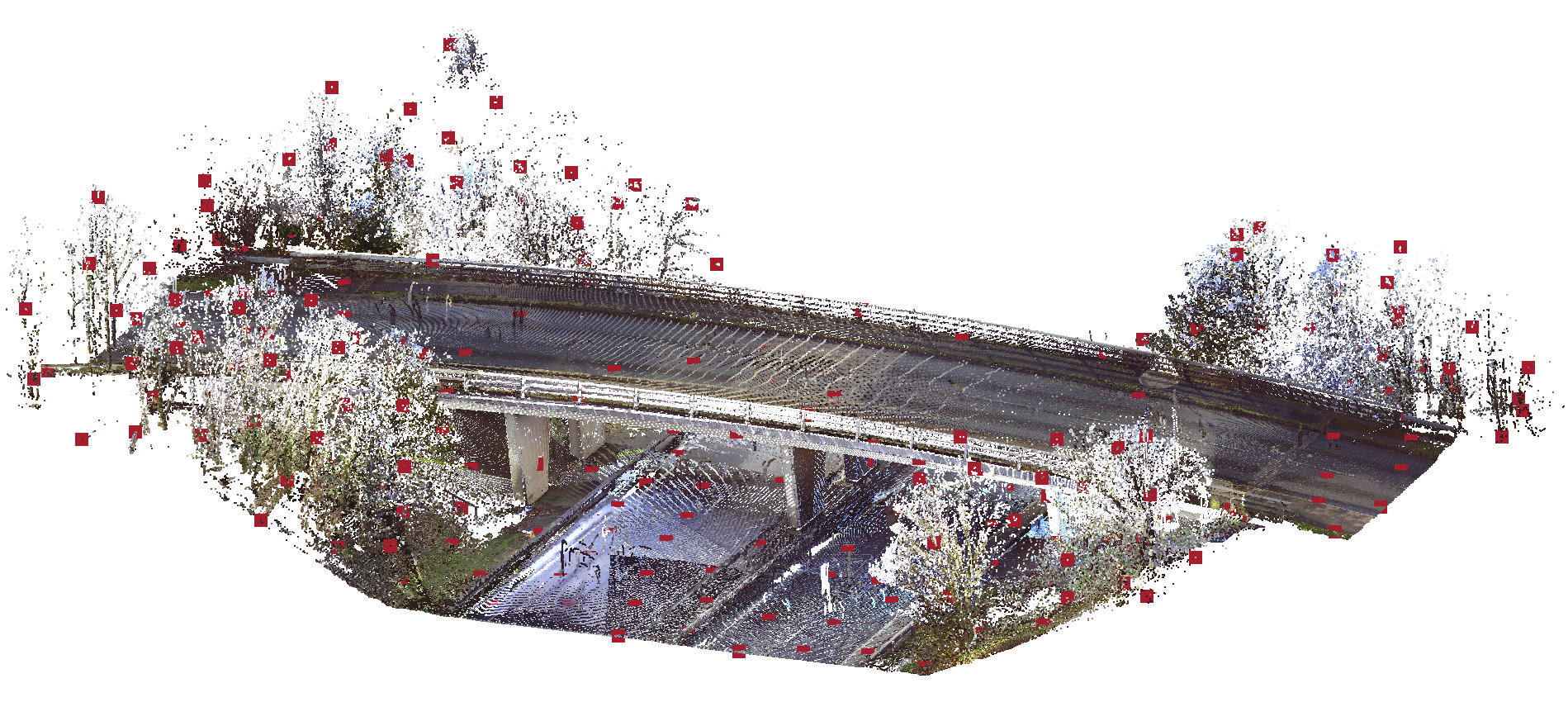}
        \caption{Full point cloud with randomly selected center positions $\mathbf{x}_c$}
    \end{subfigure}
    \hfill
    \begin{subfigure}{0.45\textwidth}
        \includegraphics[width=\textwidth]{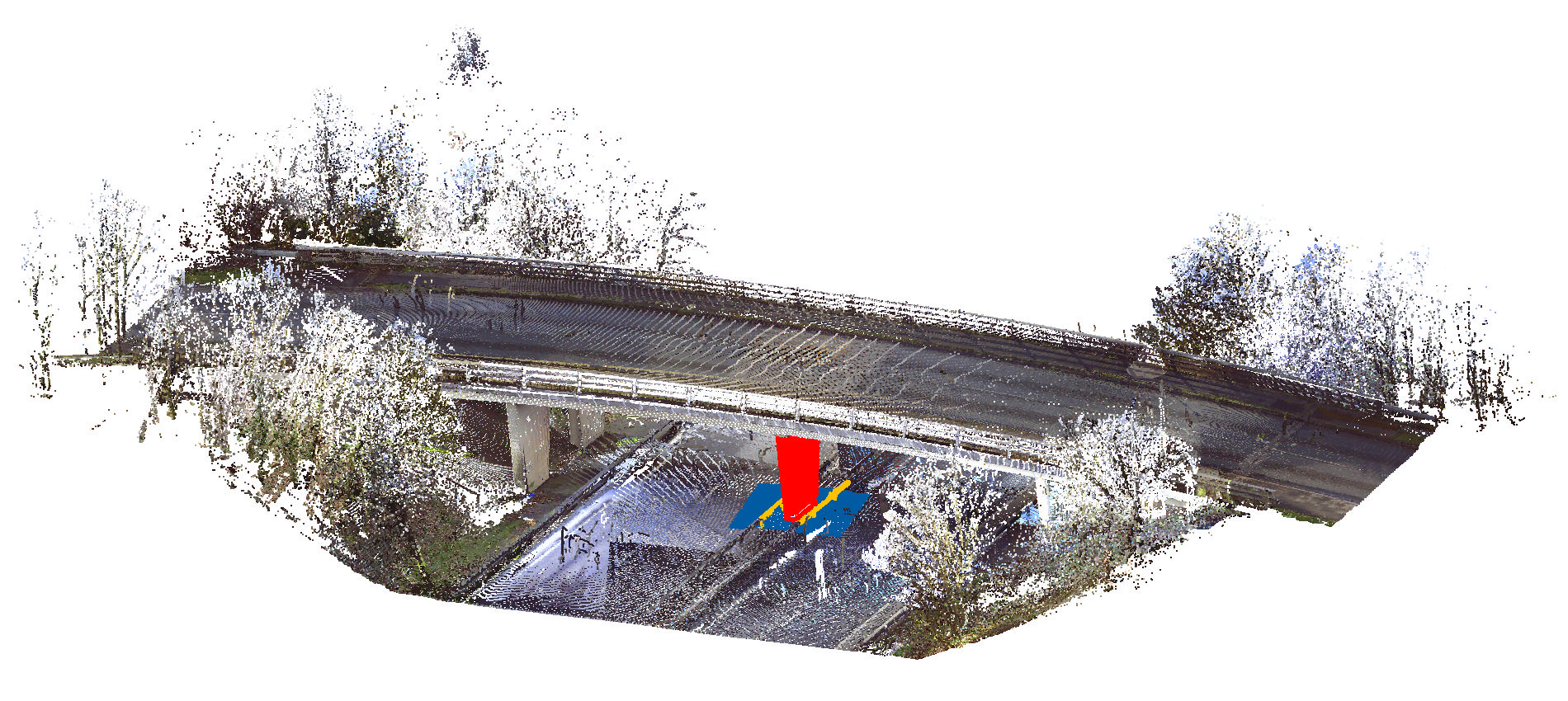}
        \caption{Full point cloud with a highlighted subcloud using annotation.}
    \end{subfigure}
    \caption[]{Visualization of point clouds showing a colored bridge and the randomly calculated center points for subcloud calculation.}
    \label{fig:suppl_example_sub_clouds}
\end{figure}

Similarly to other 3D datasets, the scenes within ours are of a size that is too large to be semantically segmented in a single operation. Consequently, we adopt the approach proposed by \cite{KPConv}, whereby the entire scene is split into subclouds. This is achieved by selecting a random position $\mathbf{x}_c$ and creating a 3D bounding box with the bounding $\mathbf{b}$ around it to create a subcloud $ \mathcal{X}^p = \{ \mathbf{p}_i \in P \,|\, |p_i^x - x_c^x| \leq b^x, \, |p_i^y - x_c^y| \leq b^y, \, |p_i^z - x_c^z| \leq b^z \} $. In our particular case, we have selected $b^x = b^y = b^z = 3$; however, it would be reasonable to consider alternative options with the objective of obtaining a more elevated perspective along the z-dimension, for example. The calculation for the subclouds is performed until each point within the point cloud is visible in at least one instance. To exemplify, we present a visual representation of randomly selected center points in conjunction with the complete cloud, as illustrated in Figure \ref{fig:suppl_example_sub_clouds}. Furthermore, to provide a sense of scale, we include the annotation of a single subcloud within the full scene, as shown in the same Figure. The number of subclouds calculated per bridge, as well as the mean number of points within each subcloud, were monitored throughout the training phases and are presented in Table \ref{tab:amount_of_subclouds}.

\begin{table}
\small
\centering
\begin{tabular}{lrr}
    \hline
    Bridge & Average subclouds & Average pts \\
    \hline
    bridge\_2\_cb & 237.2 & \numprint{101375} \\
    bridge\_3\_cb & 207.7 & \numprint{126729} \\
    bridge\_4\_cb & 237.1 & \numprint{89418} \\
    bridge\_5\_cb & 251.2 & \numprint{79383} \\
    bridge\_6\_cb & 256.4 & \numprint{83865} \\
    bridge\_8\_cb & 200 & \numprint{94727} \\
    bridge\_9\_cb & 298.7 & \numprint{78465} \\
    bridge\_11\_fr & 78 & \numprint{170413} \\
    bridge\_12\_fr & 239 & \numprint{222892} \\
    bridge\_14\_fr & 192.7 & \numprint{115061} \\
    bridge\_15\_fr & 87.1 & \numprint{114306} \\
    bridge\_16\_fr & 377.2 & \numprint{82155} \\
    bridge\_18\_fr & 299.1 & \numprint{133638} \\
    bridge\_20\_fr & 367.2 & \numprint{133673} \\
    \hline
     & 3575.6 \\
     \hline
\end{tabular}
\caption[]{The mean number of calculated subclouds for each bridge within the training set, along with the mean number of points within each subcloud (amount is calculated on original data). The mean number of subclouds used in a single epoch was found to be 3575.}
\label{tab:amount_of_subclouds}
\end{table}

\subsection{Comparison and Results}


 Two evaluations have been considered in this proposed benchmark. The initial evaluation comprises the assessment of TLS records from the defined test split, as outlined in Section \ref{ch:data}. The objective of this evaluation is to compare the performance among the implemented architectures and generally identify the challenging classes of our dataset. In the next evaluation, we only consider the point clouds from our chosen test split that were recorded for the same bridge with both the TLS and MLS scanners (test indices 13, 17 and 19 for clouds fr\_rtc and fr\_blk). We report the results per sensor. This evaluation enables the investigation of the domain gap introduced by the change in the sensor at test time, since the only difference, among the inputs seen by the models, is the change in the source of the data. The evaluation is conducted by employing the mean intersection over union (mIoU) metric, which is the most prevalent metric in the evaluation of semantic segmentation issues.

\begin{table*}
\footnotesize
\centering
\begin{tabular}{ccccccccccccc}\hline
    \rotatebox{70}{Data} & \rotatebox{70}{Model} & \rotatebox{70}{Unlabeled} & \rotatebox{70}{Underground} & \rotatebox{70}{High Vegetation} & \rotatebox{70}{Abutment} & \rotatebox{70}{Superstructure} & \rotatebox{70}{Deck} & \rotatebox{70}{Railing} & \rotatebox{70}{Traffic Sign} & \rotatebox{70}{Pillar} & \rotatebox{70}{mIoU} \\
    \hline
    \multirow{3}{*}{TLS} & UNet3D & 0.888 & \textbf{0.786} & 0.877 & 0.508 & 0.894 & 0.88 & 0.759 & \textbf{0.186} & \textbf{0.583} & \textbf{0.707} \\
     & KPConv & \textbf{0.895} & 0.784 & 0.872 & \textbf{0.599} & \textbf{0.898} & \textbf{0.89} & \textbf{0.796} & 0.064 & 0.544 & 0.705\\
     & PTv2 & 0.812 & 0.758 & \textbf{0.884} & 0.502 & 0.705 & 0.756 & 0.752 & 0.158 & 0.389 & 0.635 \\
    \hline
    \multirow{3}{*}{TLS} & UNet3D & 0.691 & 0.722 & 0.853 & 0.687 & \textbf{0.894} & \textbf{0.882} & 0.7 & \textbf{0.196} & 0.514 & 0.682 \\
    & KPConv & \textbf{0.763} & 0.714 & 0.85 & \textbf{0.796} & 0.892 & 0.853 & \textbf{0.711} & 0.105 & \textbf{0.697} & \textbf{0.709} \\
     & PTv2 & 0.661 & 0.727 & 0.857 & 0.663 & 0.884 & 0.869 & 0.672 & 0.083 & 0.382 & 0.644 \\
    \hline
    \multirow{3}{*}{MLS} & UNet3D & 0.505 & \textbf{0.733} & 0.856 & 0.578 & 0.73 & 0.618 & 0.625 & 0.13 & 0.387 & 0.574 \\
     & KPConv & 0.613 & 0.731 & \textbf{0.864} & 0.642 & 0.737 & 0.631 & 0.636 & 0.069 & 0.431 & 0.595 \\
     & PTv2 & 0.526 & 0.732 & 0.852 & 0.547 & 0.737 & 0.631 & 0.607 & 0.161 & 0.377 & 0.575 \\
    \hline
 \multirow{3}{*}{$\Delta$}  & UNet3D & -0.186 & 0.011 & \textbf{0.003} & \textbf{-0.109} & -0.164 & -0.264 & -0.075 & -0.066 & -0.127 & -0.108 \\
     & KPConv & -0.15 & 0.017 & 0.014 & -0.154 & -0.155 & \textbf{-0.222} & -0.075 & \textbf{-0.036} & -0.266 & -0.114 \\
     & PTv2 & \textbf{-0.135} & \textbf{0.005} & -0.005 & -0.116 & \textbf{-0.147} & -0.238 & \textbf{-0.065} & 0.078 & \textbf{-0.005} & \textbf{-0.069} \\
    \hline
 \\
\end{tabular}
\caption{Evaluations results. The initial block presents the IoU for each class on the test set, utilizing solely TLS data. The subsequent two blocks illustrate the outcomes of the trained networks when applied to the identical data captured by disparate sensors (see Table \ref{tab:data_splits}). The first of these employs the TLS test data. The second utilizes the MLS test data. The final block is the nominal difference between the MLS and TLS results (from third and second block, respectively). In bold best results for each evaluation considering first, second and third, and last block for the comparison.}
\label{tab:results_ai}
\end{table*}

\begin{figure*}[ht]
    \centering
    \begin{subfigure}{0.3\textwidth}
        \includegraphics[width=\textwidth]{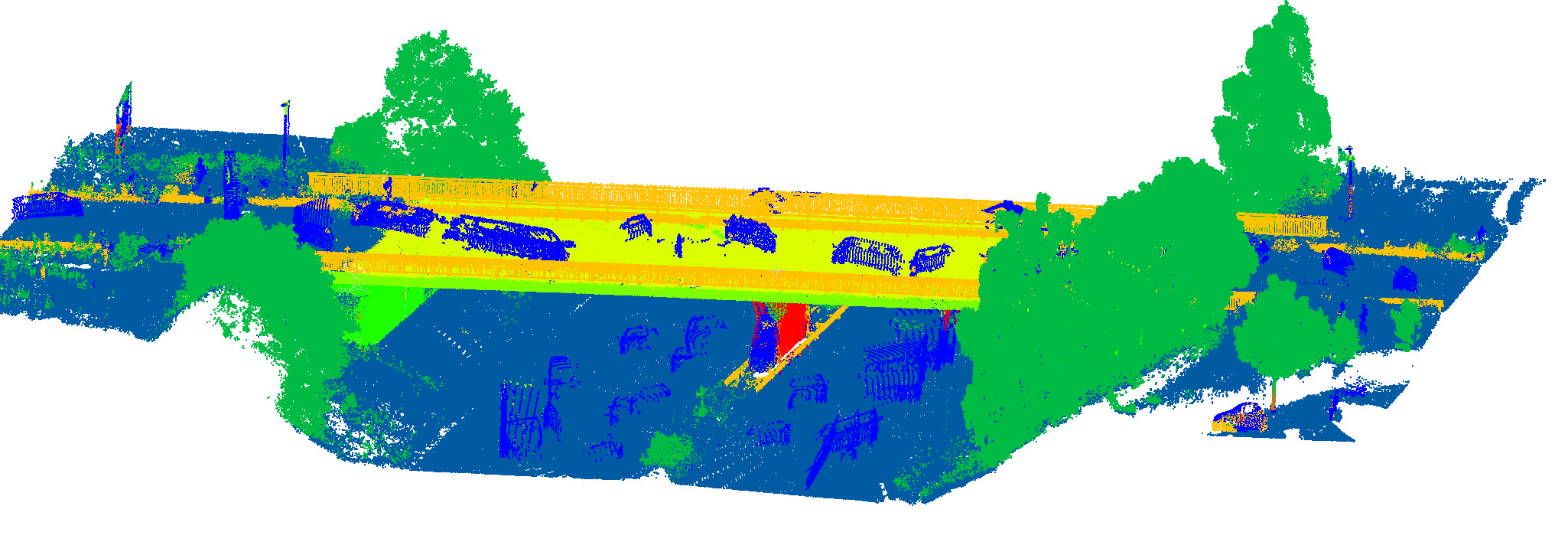}
    \end{subfigure}
    \begin{subfigure}{0.3\textwidth}
        \includegraphics[width=\textwidth]{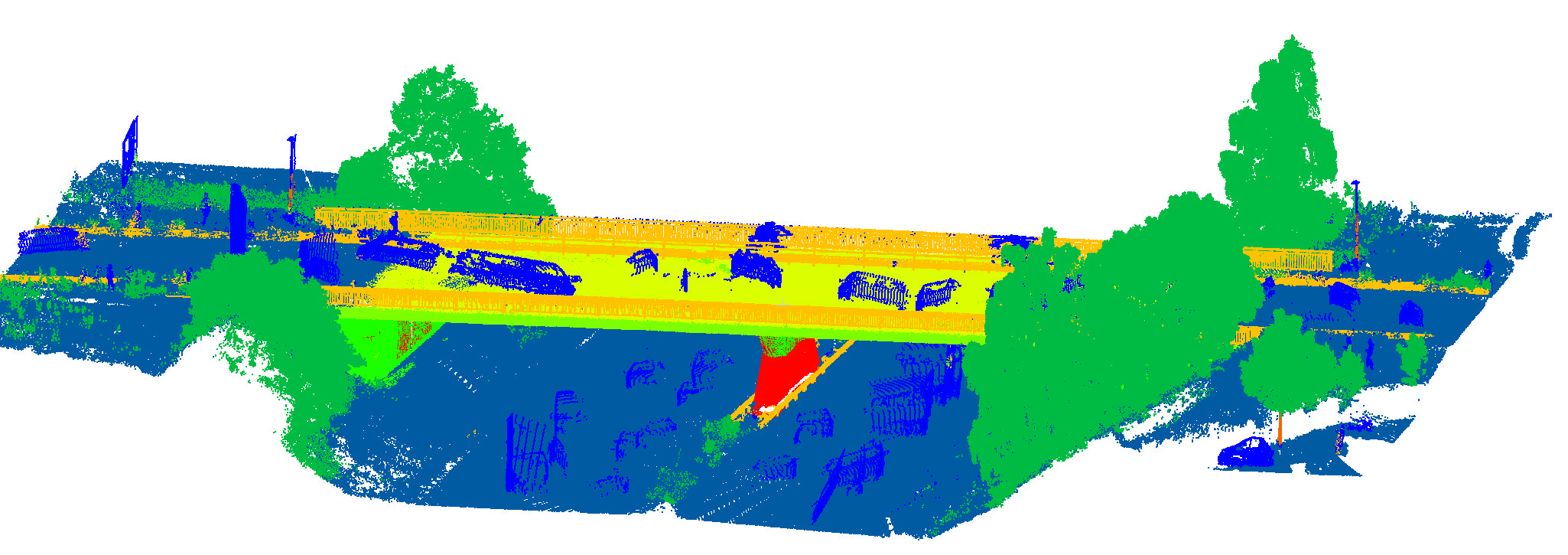}
    \end{subfigure}
    \begin{subfigure}{0.3\textwidth}
        \includegraphics[width=\textwidth]{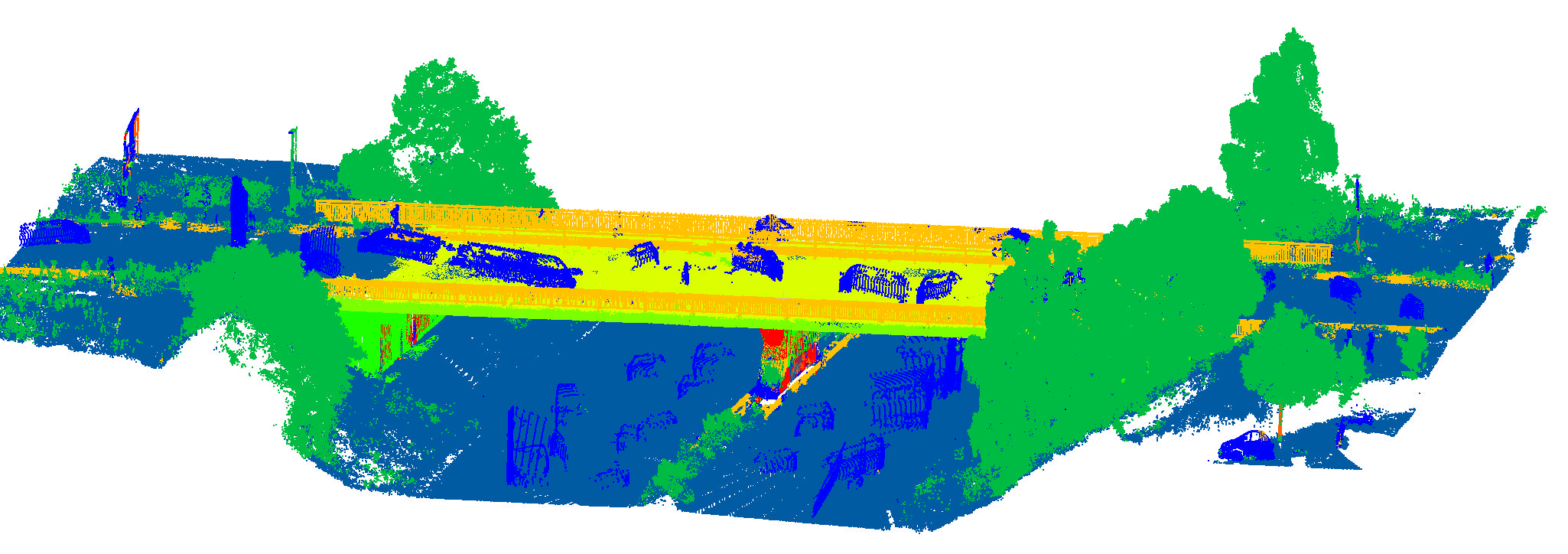}
    \end{subfigure}
    \begin{subfigure}{0.3\textwidth}
        \includegraphics[width=\textwidth]{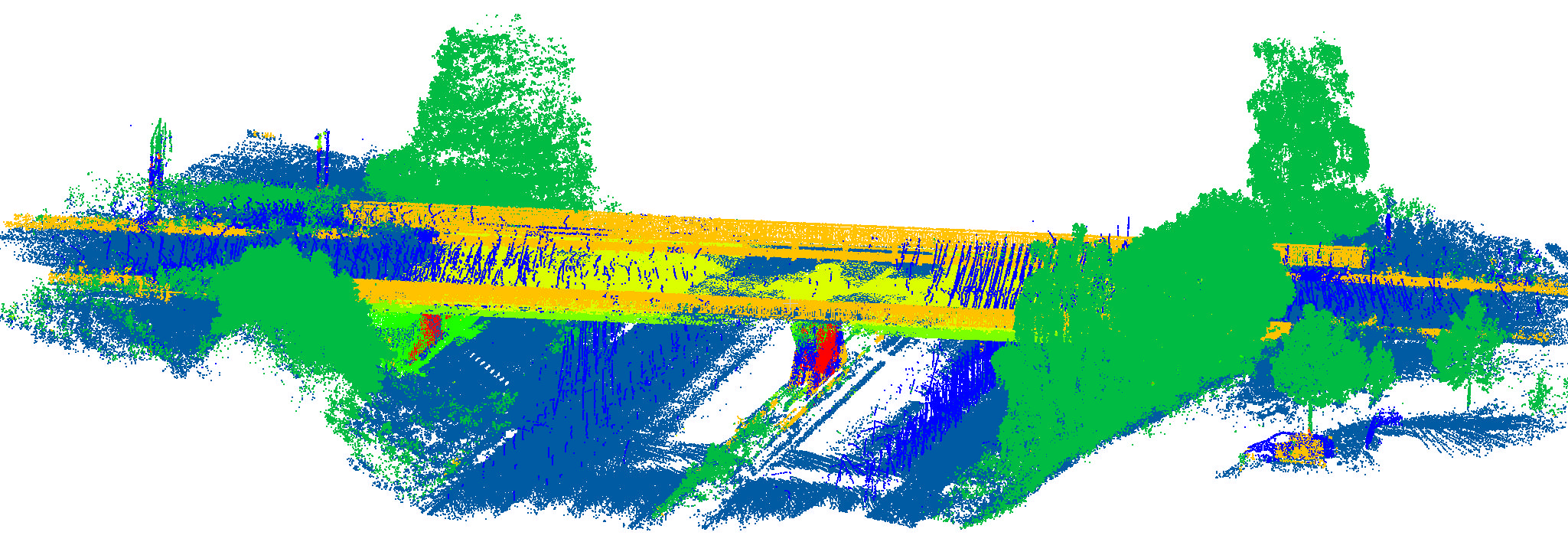}
    \end{subfigure}
    \begin{subfigure}{0.3\textwidth}
        \includegraphics[width=\textwidth]{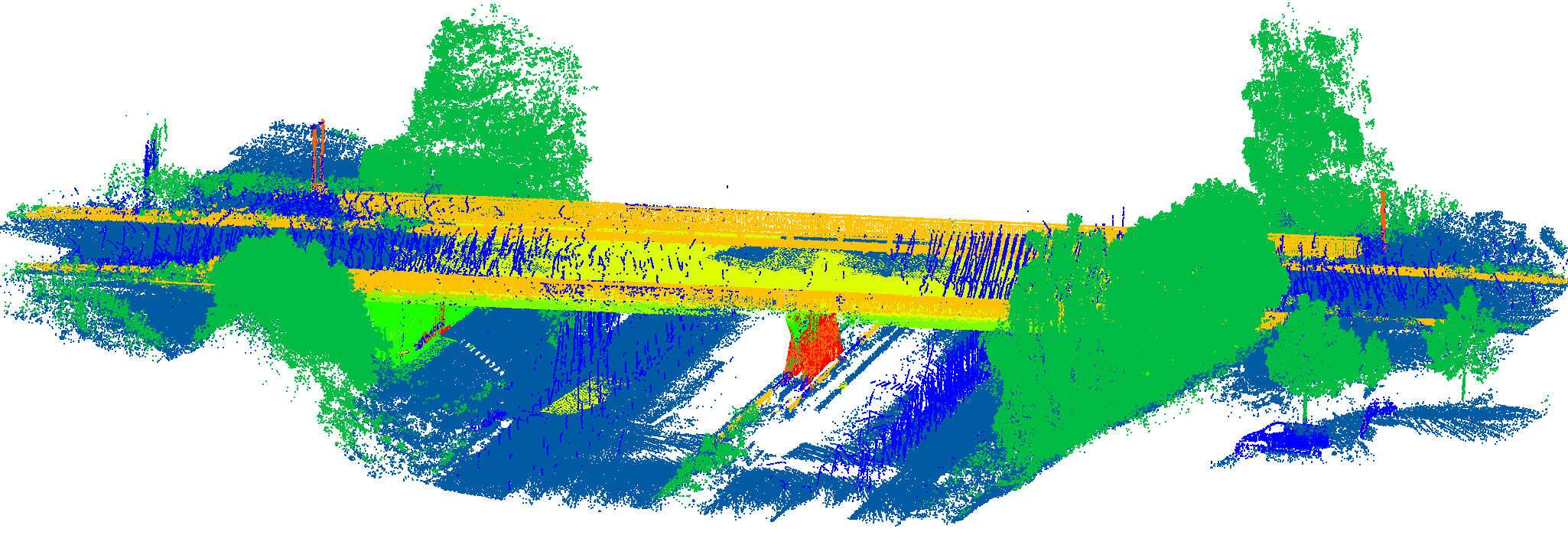}
    \end{subfigure}
    \begin{subfigure}{0.3\textwidth}
        \includegraphics[width=\textwidth]{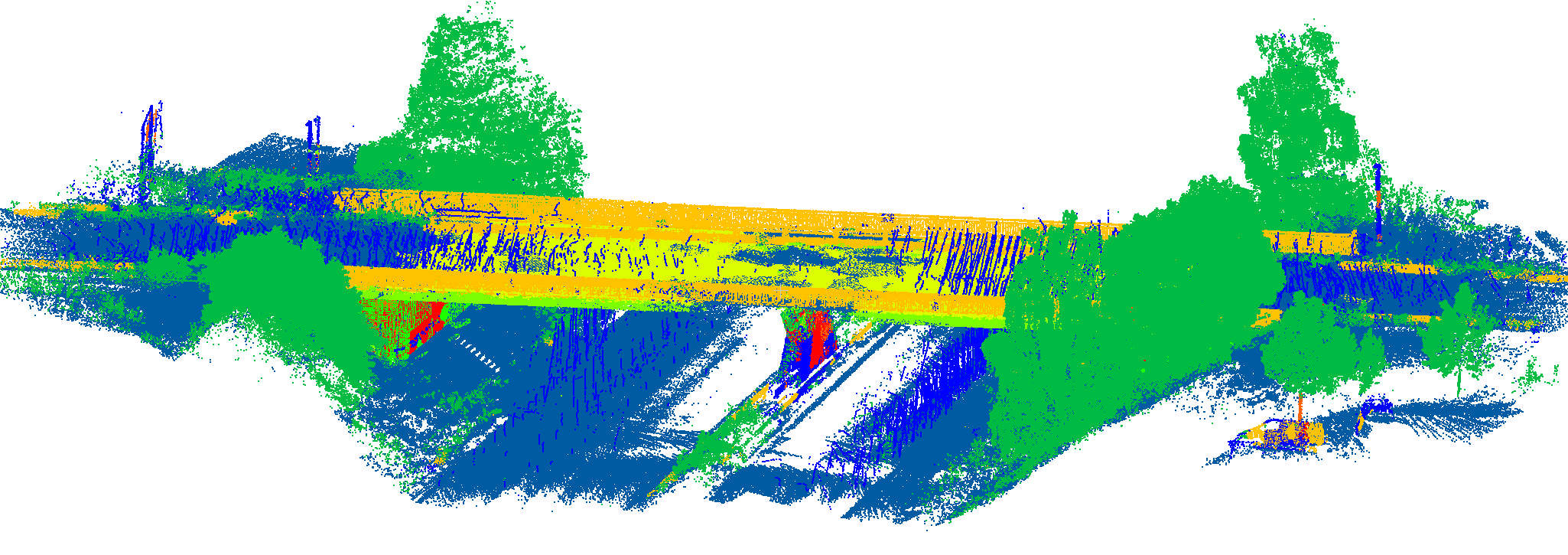}
    \end{subfigure}
    \begin{subfigure}{\textwidth}
    {
        \begin{tikzpicture}
        \draw [fill=white,draw=none] (0,.5) rectangle +(16,1.1);
        
        \foreach \x/\y/\z in {0/c_abutment/Abutment,
                              1.8/c_deck/Deck,
                              3.1/c_railing/Railing,
                              4.6/c_superstructure/Superstructure,
                              7.0/c_pillar/Pillar,
                              8.2/c_unlabeled/Unlabeled,
                              10.1/c_tsign/Traffic Sign,
                              12.1/c_ground/Ground,
                              13.6/c_hvegetation/High Vegetation}
            {
            \fill[color=\y] (\x+.1, 1) rectangle +(0.3, 0.3);
            \node [font=\footnotesize, anchor=west] at (\x+.3, 1.1) {\z}; 
            }
                
        \end{tikzpicture}
    }
    \end{subfigure}   
    \caption[]{Visualization of example predictions. The initial column depicts the outcomes obtained with the UNet3D, the subsequent column illustrates the results achieved with KPConv, and the final column presents the outcomes attained with PTv2. The top row of point clouds was captured using the stationary sensor, while the lower row was captured by the mobile scanner. Zoom in for a better view.}
    \label{fig:example_predictions}
\end{figure*}

An illustration of some example results is included in Figure \ref{fig:example_predictions}. The results of our general test and the sensor-conditioned tests are presented in Table \ref{tab:results_ai}, where the first block of results corresponds to our general evaluation whereas the next two blocks correspond each to the TLS and the MLS evaluations of the same set of bridges, respectively. The last block is the difference of the MLS and TLS results from the two previous blocks. Following, we discuss our findings regarding the observed results.


Overall, all networks achieved adequate results, that is, results larger than 0.5 IoU, in most of the classes. The most challenging classes for all the models are the "Traffic Sign" class, followed by the "Pillar" and the "Abutment" classes. The UNet3D and the KPConv models exhibited similar performance except for the "Traffic Sign" class, where the UNet3D nearly tripled the performance shown by the KPConv model. Among the five classes belonging to the bridge structure, KPConv shows the best result except for the "Pillar" class, where the UNet3D model leads by 3.9\% and 19.4\% points over KPConv and PTv2, respectively. Comparably, the PTv2 model, attained the lowest mIoU result. Obtaining the best result among the three models only for the vegetation class and not with a large margin with respect to the next best model (i.e., 0.7\% points difference). This is a noteworthy outcome, considering that the reported performance of this same model for the S3DIS indoor dataset \cite{S3DIS}, for instance, is superior to that of the rest of the compared models.

The analysis of the performance of the transferred data yielded different results. All three models tested showed a drop in performance. The KPConv model shows the largest performance drop of 11.4\% points, closely followed by the UNet3D model with a drop of 10.8\% points. In contrast, PTv2 shows a relatively smaller performance drop of 6.9\%. It is noteworthy that this evaluation shows the domain gap effect overall, since the models and the target bridges are the same in both test setups, the only change being the source of the input data. The least affected classes are "Underground" and "High Vegetation", which barely show changes in the prediction metric. Regarding the classes related to the bridge structure, the most affected case overall is the prediction using the KPConv model for the "Pillar" class with a drop of 26.60\% points. On the other hand, the least affected case in the same group is that of the "Pillar", but for the PTv2 model, with a drop of only 0.50\% points. For the three models, on average, the degradation of performance mainly affects the "Top Surface", "Superstructure" and "Abutment" classes, in that order. The "Traffic Sign" class predicted by the PTv2 model case shows the greatest improvement when compared to the results obtained using the TLS data, with an improvement of 7.80\% points.

\section{Discussion}\label{ch:discussion}
The findings of this study demonstrate that the data can be utilised to train various neural networks, thereby achieving satisfactory performance in all cases. All classes can be effectively learned, with the exception of the "traffic sign" class, which is not a significant concern due to its frequent coverage in other datasets and its irrelevance to the bridges themselves. It should be noted that the neural network architecture itself was not optimized for the purpose of segmenting bridges. However, an increase in performance can be expected by using different hyperparameters, such as the initial voxel size or the size of subclouds, for instance.

An analysis of the performance of the models on transferred data revealed a consistent decline across all three networks, thereby highlighting the impact of domain adaptation challenges. The KPConv model exhibited the most significant drop, with a decrease of 11.4\% points, followed by UNet3D with 10.8\% points. Conversely, PTv2 demonstrated a comparatively minor decrease of 6.9\%, indicating its potential resilience to variations in input data sources. This observed degradation in performance underscores the domain gap effect caused by the sensor, as the models were evaluated on the same target bridges, with the only variable being the source of input data.

It should be noted that the four additional bridges recorded with an MLS were not utilized in the present study. However, it is important to acknowledge the potential value of these bridges for future adaptation techniques in further research.

There are several limitations that require discussion. Primarily, the type of bridge is of significance. All of the bridges presented are of the same type, which could potentially result in poor performance when applying the trained networks to an arch bridge, for instance, due to the learned bias towards this specific bridge type. Beyond bridge type, all the bridges in this dataset were collected from locations in the UK and Germany. Bridges in other regions may differ due to variations in local design regulations and construction practices, which could impact model generalization. Lastly, this study explores the influence of the domain gap using only two sensor types in the dataset of bridges. To gain a more comprehensive understanding of the domain gap, future research should be considered using a wider variety of sensors and data collected from more diverse environments.

\section{Conclusion}\label{ch:conclusion}
This study introduces two key contributions. Firstly, a new large dataset was created for the purpose of 3D semantic segmentation of bridges. To the best of our knowledge, this is the first annotated point cloud dataset of bridges, containing 20 bridges collected from the UK and Germany. Notably, it includes classes such as abutments and pillars, which are absent in existing datasets. This dataset serves as a valuable benchmark for evaluating newly proposed deep learning models for semantic segmentation, contributing to advancements in infrastructure digitalization. Secondly, the impact of the 3D sensor on the performance of the trained neural networks using different sorts of representations was investigated. The findings measure the impact of the domain gap introduced by sensor change. The setup of our dataset allows the usage and validation of different domain adaptation approaches to further research relevant to the objective.

\section*{Acknowledgment}

The work has been funded by the BMDV as part of the mFUND project {``Partially automated creation of object-based inventory models using multi-data fusion of multimodal data streams and existing inventory data - mdfBIM+``} (FKZ: 19FS2021B).

\bibliographystyle{elsarticle-num} 
\bibliography{chapter/refs}



\end{document}